\documentclass[runningheads]{llncs}
\usepackage{times}
\usepackage{url}
\usepackage{latexsym}
\usepackage{multirow}
\usepackage{graphicx}
\usepackage{soul}
\usepackage{amsmath}
\usepackage{bm}
\usepackage{microtype}
\usepackage{natbib}
\usepackage{color}
\usepackage{booktabs}
\usepackage[utf8]{inputenc}
\usepackage{CJKutf8}
\usepackage[justification=centering]{subfig}
\usepackage[hidelinks]{hyperref}
\usepackage[nameinlink,capitalize]{cleveref}
\usepackage{authblk}

\DeclareMathOperator*{\argmax}{arg\,max}

\date{}

\begin{document} 
    \title{How Much Does Tokenization Affect Neural Machine Translation?}
    \author{Miguel Domingo\inst{1} \and Mercedes Garc\'ia-Mart\'inez\inst{2} \and Alexandre Helle\inst{2} \and Francisco Casacuberta\inst{1} \and Manuel Herranz\inst{2}}
    \authorrunning{M. Domingo, M. Garc\'ia-Mart\'inez, A. Helle, F. Casacuberta and M. Herranz}
    \institute{Pattern Recognition and Human Language Technology Research Center \\ Universitat Polit{\`e}cnica de Val{\`e}ncia \\ Camino de Vera s/n, 46022 Valencia, Spain \\ \email{\{midobal, fcn\}@prhlt.upv.es} \and Pangeanic / B.I Europa \\ PangeaMT Technologies Division \\ Valencia, Spain \\ \email{\{m.garcia, a.helle, m.herranz\}@pangeanic.com}}
    
    \maketitle
    
    \begin{abstract}
        Tokenization or segmentation is a wide concept that covers simple processes such as separating punctuation from words, or more sophisticated processes such as applying morphological knowledge. 
        Neural Machine Translation (NMT) requires a limited-size vocabulary for computational cost and enough examples to estimate word embeddings.
        Separating punctuation and splitting tokens into words or subwords has proven to be helpful to reduce vocabulary and increase the number of examples of each word, improving the translation quality. 
        Tokenization is more challenging when dealing with languages with no separator between words. In order to assess the impact of the tokenization in the quality of the final translation on NMT, we experimented on five tokenizers over ten language pairs. We reached the conclusion that the tokenization significantly affects the final translation quality and that the best tokenizer differs for different language pairs. 
    \end{abstract}
    
    \section{Introduction}
    
    Segmentation is an essential process that has been extensively studied in literature~\citep{niessen04,goldwater05,dyer09,nguyen10}. It covers simple processes such as separating punctuation from words (tokenization), splitting words in subparts based on their frequency or more sophisticated processes such as applying morphological knowledge. In this work, we use tokenization referring to separating punctuation and splitting tokens into words or subwords.
    
    Tokenizing words has proven to be helpful to reduce vocabulary and increase the number of examples of each word. It is extremely important for languages in which there is no separation between words and, therefore, a single token corresponds to more than one word. The way in which tokens are split can greatly change the meaning of the sentence. For example, the Japanese word \begin{CJK}{UTF8}{min}警\end{CJK} means \emph{admonish}, and \begin{CJK}{UTF8}{min}察\end{CJK} means \emph{observe}. However, together they form the word \emph{police} (\begin{CJK}{UTF8}{min}警察\end{CJK}). Therefore, a correct tokenization can help to improve translation quality.
    
    In this study, we aim to find the impact of tokenization on the quality of the final translation. To do so, we experimented with five tokenizers over ten language pairs. To the best of our knowledge, this is the first work in which an exhaustive comparison between tokenizers has been run for NMT. We include tokenizers based on morphology that could guide the splitting of the words~\citep{Pinnis2017NeuralMT}.
    
    Some previous works include studying the effect of word-level preprocessing for Arabic on Statistical Machine Translation (SMT). A comparison of several segmenters for Chinese on SMT was done by \citet{ZhaoZH}. \citet{W17-4706} compared morphological segmenters for German in NMT. Finally, \citet{taku18b} compared his statistical word segmenter with other well-known Japanese morphological segmenters, reaching the conclusions that statistical segmenters worked better than morphological ones.
    
    Our main contributions are as follows:
    
    \begin{itemize}
        \item First study of tokenizers for neural machine translation.
        \item Experimentation with five different tokenizers over ten language pairs.
    \end{itemize}
    
    The rest of this document is structured as follows: \cref{SecNMT} introduces the neural machine translation system used in this work. After that, in \cref{SecWS}, we present the tokenizers applied for comparison purposes. Then, in \cref{SecExperiments}, we describe the experimental framework, whose results are presented and discussed in \cref{SecResults}. \cref{SecAnalysis} shows some translation examples of the results. Finally, in \cref{SecConclusion}, conclusions are drawn.
    
    \section{Neural Machine Translation}
    \label{SecNMT}
    Given a source sentence $x_1^J=x_1,\dots,x_J$ of length $J$, NMT aims to find the best translated sentence $\hat{y}_1^{\hat{I}}=\hat{y}_1,\dots,\hat{y}_{\hat{I}}$ of length $\hat{I}$:
    
    \begin{equation}
    \hat{y}_1^{\hat{I}} = \argmax_{I,y_1^I} Pr(y_1^I \mid x_1^J)
    \end{equation}
    
    \noindent where the conditional translation probability is modelled as:
    
    \begin{equation}
    Pr(y_1^I \mid x_1^J) = \prod_{i=1}^{I} Pr(y_i \mid y_1^{i-1},x_1^J)
    \end{equation}
    
    NMT frequently relies on a Recurrent Neural Network (RNN) encoder-decoder framework. The source sentence is projected into a distributed representation at the encoding step. Then, the decoder generates, at the decoding step, its translation word by word~\citep{Sutskever14}.
    
    The input of the system is a word sequence in the source language. Each word is projected linearly to a fixed-size real-valued vector through an embedding matrix. Then, these word embeddings are fed into a bidirectional~\citep{schuster97} Long Short-Term Memory (LSTM)~\citep{hochreiter97} network. As a result, a sequence of annotations is produced by concatenating the hidden states from the forward and backward layers.
    
    An attention mechanism~\citep{Bahdanau15} allows the decoder to focus on parts of the input sequence, computing a weighted mean of annotated sequences. A soft alignment model computes these weights, weighting each annotation with the previous decoding state.
    
    Another LSTM network is used for the decoder. This network is conditioned by the representation computed by the attention model and the last generated word. Finally, a distribution over the target language vocabulary is computed by the deep output layer~\citep{pascanu13}.
    
    The model is trained by applying stochastic gradient descent jointly to maximize the log-likelihood over a bilingual parallel corpus. At decoding time, the model approximates the most likely target sentence with beam-search~\citep{Sutskever14}.
    
    \section{Tokenizers}
    \label{SecWS}
    In this section, we present the tokenizers we employed in order to assess their impact on the quality of the final translation.
    
    \begin{description}
        \item[SentencePiece\footnotemark{\normalfont :}]\footnotetext{\url{https://github.com/google/sentencepiece}} an unsupervised text tokenizer and detokenizer mainly for Neural Network-based text generation systems where the vocabulary size is predetermined prior to the neural model training. It can be used for any language, but its models need to be trained for each of them. To do so, we used the unigram~\citep{kudo18} mode and a vocabulary size of 32000 over each corpora's training partition. \cref{FigSentencePiece} shows an example of tokenizing a sentence using \emph{SentencePiece}.
        \item[Mecab\footnotemark{\normalfont :}]\footnotetext{\url{http://taku910.github.io/mecab/}} an open source morphological analysis engine for Japanese, based on conditional random fields. It extracts morphological and syntactical information from sentences and splits tokens into words. \cref{FigMecab} shows an example of tokenizing a sentence using \emph{Mecab}.
        \item[Stanford Word Segmenter]\citep{Tseng2005}: a Chinese word segmenter based on conditional random fields. Using a set of morphological and character reduplication features, it is able to split Chinese tokens into words. In this work, we use the toolkit's CTB scheme. \cref{FigStanford} shows an example of tokenizing a sentence using \emph{Stanford Word Segmenter}.    
        \item[OpenNMT tokenizer]\citep{OpenNMT}: the tokenizer included with the \emph{OpenNMT} toolkit. It normalizes characters (e.g., quotes Unicode variants) and separates punctuation from words. It can be used with any language. \cref{FigOpenNMT} shows an example of tokenizing a sentence using \emph{OpenNMT tokenizer}.    
        \item[Moses tokenizer]\citep{Moses}: the tokenizer included with the \emph{Moses} toolkit. It separates punctuation from word {\textemdash}preserving special tokens such as URL or dates{\textemdash} and normalizes characters (e.g., quotes Unicode variants). It can be used with any language. \cref{FigMoses} shows an example of tokenizing a sentence using \emph{Moses tokenizer}.
    \end{description}
    
    \begin{figure}
        \begin{flushleft}
            Original: \hspace{3.5mm} \emph{In a browser window (Internet Explorer or Firefox) browse to www.dellconnect.com.} \\
            Segmented: \emph{\_In \_a \_browser \_window \_( Internet \_Explorer \_or \_Firefox ) \_browse \_to \_www . dell connect . com .}
        \end{flushleft}
        \subfloat[\label{FigSentencePiece} Example of a sentence tokenized using \emph{SentencePiece}. \emph{\_} indicates the start of a word in the original sentence. The tokenization has split punctuation and transformed the url into several words.]{\hspace{\linewidth}}
        \begin{flushleft}
            $ $ \\
            $ $ \\
            Original:\hspace{3.5mm} \emph{\begin{CJK}{UTF8}{min}ブラウザウィンドウ(Internet ExplorerまたはFirefox)で、www.dellconnect.comにアクセスします。\end{CJK}}\\
            Segmented: \emph{\begin{CJK}{UTF8}{min}ブラウザウィンドウ ( Internet Explorer または Firefox ) で 、 www . dellconnect . com に アクセス し ま す 。\end{CJK}}
        \end{flushleft}
        \subfloat[\label{FigMecab} Example of a sentence tokenized using \emph{Mecab}. In the original sentence, the only spaces were written to separate foreign words (\emph{Internet Explorer}). The tokenization has added spaces between Japanese words, split the punctuation and transformed the url into several words.]{\hspace{\linewidth}}
        \begin{flushleft}
            $ $ \\
            $ $ \\
            Original: \hspace{3.5mm} \emph{\begin{CJK}{UTF8}{min}转到 http://www.kace.com/trial，然后单击“下载 K1000 试用版”，将压缩的 OVF（开放虚拟化格式）文件下载到 vSphere 系统。\end{CJK}}\\
            Segmented: \emph{\begin{CJK}{UTF8}{min}转到 http : //www.kace.com/trial ， 然后 单击 “ 下载 K1000 试用版 ” ， 将 压缩 的 OVF （ 开放 虚拟化 格 式 ） 文件 下载 到 vSphere 系统 。\end{CJK}}
        \end{flushleft} 
        \subfloat[\label{FigStanford}Example of a sentence tokenized using \emph{Stanford Word Segmenter}. The original sentence only contained spaces to separate foreign words (e.g., \emph{vSphere}). The tokenization has added spaces between the Chinese words, split the punctuation, and separated the \emph{http:} from the url.]{\hspace{\linewidth}}
        \begin{flushleft}
            $ $ \\
            $ $ \\
            Original: \hspace{3.5mm} \emph{In a browser window (Internet Explorer or Firefox) browse to www.dellconnect.com.} \\
            Segmented: \emph{In a browser window ( Internet Explorer or Firefox ) browse to www . dellconnect . com .}
        \end{flushleft}
        \subfloat[\label{FigOpenNMT} Example of a sentence tokenized using \emph{OpenNMT tokenizer}. The tokenization has split punctuation and transformed the url into several words.]{\hspace{\linewidth}}
        \begin{flushleft}
            $ $ \\
            $ $ \\
            Original: \hspace{3.5mm} \emph{In a browser window (Internet Explorer or Firefox) browse to www.dellconnect.com.} \\
            Segmented: \emph{In a browser window ( Internet Explorer or Firefox ) browse to www.dellconnect.com.}
        \end{flushleft}
        \subfloat[\label{FigMoses} Example of a sentence tokenized using \emph{Moses tokenizer}. The tokenization has split the punctuation, without modifying the url.]{\hspace{\linewidth}}
        \caption{Examples of segmenting sentences with each word segmenter.}
    \end{figure}
    
    \section{Experimental Framework}
    \label{SecExperiments}
    In this section, we describe the corpora, systems and metrics used in order to asses our proposal.
    
    \subsection{Corpora}
    The corpora selected for our experimental session was extracted from translation memories from the translation industry.  The files are the result of professional translation tasks demanded by real clients. The general domain is technical (see \cref{ta:domains} for the specific content of each language pair), which is harder for NMT than other general domains such as news. Unlike in other domains, in technical domains certain words correspond to specific terms and have a different translation to their most frequent one: e.g., \emph{rear arm} translates into German as \emph{hinterer Arm}. However, in this domain, it should be translated as \emph{hinterer Querlenker}. In order to increase language diversity, we selected the following language-pairs: Japanese{\textendash}English, Russian{\textendash}English, Chinese{\textendash}English, German{\textendash}English, and Arabic{\textendash}English. \cref{tabCorp} shows the corpora statistics.
    
    \begin{table}[!ht]
        \centering
        \begin{tabular}{l|c c c c c}
            \toprule
            \multirow{2}{*}{\textbf{Specific Domain}} & \multicolumn{5}{c}{\textbf{Language}} \\
            & Ja{\textendash}En & Ru{\textendash}En & Zh{\textendash}En & De{\textendash}En & Ar{\textendash}En\\
            \hline
            Computer Software - Instructions for use & & X & & & X \\
            \hline
            Medical Equipment and Supplies & X & X & X & X & X \\
            \hline
            Consumer Electronics & X & & X & X & X \\
            \hline
            Industrial Electronics & & X & & X & \\
            \hline
            Stores and Retail Distribution & X & X & X & & \\
            \hline
            Healthcare & & X & & & \\
            \bottomrule
        \end{tabular}
        \caption{Specific domains for each language pair. \emph{Ja} stands for Japanese, \emph{En} for English, \emph{Ru} for Russian, \emph{Zh} for Chinese, \emph{De} for German and \emph{Ar} for Arabic.}
        \label{ta:domains}
    \end{table}
    
    The training dataset is composed of around three million sentences in the German{\textendash}English language pair and around half a million sentences in the rest of the language pairs. Development and test datasets are composed of two thousand sentences for all the language pairs.
    
    \begin{table}[!ht]
        \centering
        \resizebox{0.9\textwidth}{!}{\begin{minipage}{\textwidth}
                \begin{tabular}{c c c c c c c}
                    \toprule
                    \multirow{2}{*}{\textbf{Partition}} & \multirow{2}{*}{\textbf{Type}} & \multicolumn{5}{c}{\textbf{Language}} \\
                    \cmidrule(lr){3-7}
                    &  & Ja{\textendash}En & Ru{\textendash}En & Zh{\textendash}En & De{\textendash}En & Ar{\textendash}En \\
                    \midrule
                    \multirow{3}{*}{Train} & Sentences & 532.0K & 496.0K & 460.8K & 2.9M & 557.0K \\
                    & Tokens & 10.0/7.3M & 7.6/7.4M & 6.7/6.4M & 35.9/39.4M & 7.3/7.8M \\
                    & Vocabulary & 41.5/111.6K & 180.9/133.3K & 82.8/102.6K & 1.1M/615.7K & 115.5/61.8K \\
                    & Tokens$_\text{BPE}$ & 10.5/8.3M & 9.8/9.5M & 7.5/7.4M & 49.8/49.0M & 8.4/8.7M \\
                    & Vocabulary$_\text{BPE}$ & 16.0/17.1K & 24.8/11.6K & 22.0/16.6K & 25.6/22.3K & 21.6/10.7K \\
                    \cmidrule(lr){1-7}
                    \multirow{3}{*}{Development} & Sentences & 2000 & 2000 & 2000 & 2000 & 2000 \\
                    & Tokens & 39.0/27.6K & 34.0/32.2K & 27.8/27.8K & 42.4/45.4K & 21.1/21.7K \\
                    & Vocabulary & 2.3/3.4K & 7.6/5.4K & 2.7/3.8K & 6.2/4.4K & 3.6/2.9K \\
                    & Tokens$_\text{BPE}$ & 42.1/31.3K & 41.2/38.5K & 29.5/31.2K & 53.7/51.0K & 23.3/24.2K \\
                    & Vocabulary$_\text{BPE}$ & 1.9/2.5K & 6.5/3.7K & 2.5/2.9K & 4.9/3.6K & 3.4/2.1K \\
                    \cmidrule(lr){1-7}
                    \multirow{3}{*}{Test} & Sentences & 2000 & 2000 & 2000 & 2000 & 2740 \\
                    & Tokens & 18.4/26.8K & 28.6/28.3K & 48.7/30.5K & 41.7/44.6K & 22.1/23.3K \\
                    & Vocabulary & 3.5/3.9K & 7.3/5.1K & 9.2/3.8K & 6.0/4.3K & 3.2/2.6K \\
                    & Tokens$_\text{BPE}$ & 39.5/30.2K & 98.7/94.4K & 32.9/35.6K & 83.9/82.8K & 34.4/32.9K \\
                    & Vocabulary$_\text{BPE}$ & 1.8/2.7K & 8.0/5.4K & 2.7/3.0K & 8.3/6.8K & 4.1/2.3K \\
                    \bottomrule
                \end{tabular}
            \end{minipage}}
            \caption{Corpora statistics. \emph{Ja} stands for Japanese, \emph{En} for English, \emph{Ru} for Russian, \emph{Zh} for Chinese, \emph{De} for German and \emph{Ar} for Arabic. \emph{Tokens$_\text{BPE}$} and \emph{Vocabulary$_\text{BPE}$} are the number of tokens and size of the vocabulary after applying BPE to the corpora. K stands for thousand and M for millions.}
            \label{tabCorp}
        \end{table}
        
        \subsection{Systems}
        NMT systems were trained with \emph{OpenNMT}~\citep{OpenNMT}. We used LSTM units taking into account the findings in \citep{britz17}. The size of the LSTM units and word embeddings were set to 1024. We used Adam~\citep{Adam} with a learning rate of 0.0002~\citep{wu16}, a beam size of 6 and a batch size of 20. We reduced the vocabulary using Byte Pair Encoding (BPE)~\citep{BPE}, training the models with a joint vocabulary of 32000 BPE units. Finally, the corpora were lowercased and, later, recased using \emph{OpenNMT}'s tools.
        
        \subsection{Evaluation metrics}
        We made use of the following well-known metrics to assess our proposal:
        
        \begin{description}
            \item[BiLingual Evaluation Understudy (BLEU)]~\citep{BLEU}: corresponds to the geometric average of the modified n-gram precision. It is multiplied by a brevity factor to penalize short sentences.
            \item[Translation Error Rate (TER)]~\citep{TER}: number of word edit operations (insertion, substitution, deletion, and swapping), normalized by the number of words in the final translation.
        \end{description}
        
        Confidence intervals ($p=0.05$) are computed for all metrics by means of bootstrap resampling~\citep{Koehn04}.
        
    \section{Results}
    \label{SecResults}
    In this section, we present the results of the experiments conducted in order to assess the impact of the tokenizer on the translation quality. \cref{tabTok} shows the experimental results.
    
    \begin{table}[!ht]
        \resizebox{0.69\textwidth}{!}{\begin{minipage}{\textwidth}
                \begin{tabular}{c c c c c c c c c c c }
                    \toprule
                    \multirow{2}{*}{\textbf{Language}} & \multicolumn{2}{c}{\textbf{SentencePiece}} & \multicolumn{2}{c}{\textbf{OpenNMT tokenizer}} & \multicolumn{2}{c}{\textbf{Moses tokenizer}} & \multicolumn{2}{c}{\textbf{Mecab}} & \multicolumn{2}{c}{\textbf{Stanford}} \\
                    \cmidrule(lr){2-3}\cmidrule(lr){4-5}\cmidrule(lr){6-7}\cmidrule(lr){8-9}\cmidrule(lr){10-11}
                    & BLEU & TER & BLEU & TER & BLEU & TER & BLEU & TER & BLEU & TER \\
                    \midrule
                    Ja{\textendash}En & $32.0 \pm 1.3$ & $51.1 \pm 1.5$ & $29.1 \pm 1.4$ & $54.7 \pm 1.4$  & $\bf{36.3 \pm 1.4}$ & $\bf{47.5 \pm 1.3}$ & $\bf{36.0 \pm 1.5}$ & $\bf{48.6\pm 1.4}$ & - & -  \\
                    En{\textendash}Ja & $26.5 \pm 1.4$ & $62.5 \pm 1.9$ & $25.0 \pm 4.4$ & $89.9 \pm 4.1$  & $33.6 \pm 2.3$ & $61.0 \pm 2.5$ & $\bf{45.8 \pm 1.3}$ & $\bf{43.7 \pm 1.3}$ & - & -  \\
                    \midrule
                    Ru{\textendash}En & $12.9 \pm 0.9$ & $72.7 \pm 1.1$ & $11.9 \pm 0.9$ & $74.9 \pm 1.3$  & $\bf{15.3 \pm 1.0}$ & $\bf{68.6 \pm 1.2}$ & - & - & - & - \\
                    En{\textendash}Ru & $12.2 \pm 0.8$ & $75.0 \pm 1.0$ & $11.3 \pm 0.9$ & $77.3 \pm 1.1$  & $\bf{16.3 \pm 1.2}$ & $\bf{70.4 \pm 1.6}$ & - & - & - & - \\
                    \midrule
                    Zh{\textendash}En & $20.5 \pm 1.1$ & $64.8 \pm 1.2$ & $23.1 \pm 1.3$ & $64.8 \pm 1.3$  & $\bf{27.5 \pm 1.3}$ & $\bf{59.8 \pm 1.2}$ & - & - & $\bf{26.0 \pm 1.3}$ & $\bf{59.3 \pm 1.2}$ \\
                    En{\textendash}Zh & $17.1 \pm 1.2$ & $71.2 \pm 1.2$ & $10.4 \pm 3.9$ & $101.1 \pm 3.1$  & $21.4 \pm 2.0$ & $65.8 \pm 1.7$ & - & - & $\bf{29.9 \pm 1.2}$ & $\bf{55.6 \pm 1.2}$ \\
                    \midrule
                    De{\textendash}En & $21.4 \pm 0.8$ & $67.8 \pm 2.1$ & $\bf{29.6 \pm 0.9}$ & $\bf{54.2 \pm 0.9}$ & $\bf{30.3 \pm 0.9}$ & $\bf{52.8 \pm 0.9}$ & - & - & - & - \\
                    En{\textendash}De & $16.1 \pm 0.7$ & $76.4 \pm 2.3$ & $\bf{22.5 \pm 0.9}$ & $\bf{65.0 \pm 1.5}$ & $\bf{23.6 \pm 0.9}$ & $\bf{62.9 \pm 1.0}$ & - & - & - & - \\
                    \midrule
                    Ar{\textendash}En & $\bf{17.9 \pm 0.8}$ & $\bf{66.9 \pm 1.}3$ & $14.8 \pm 0.8$ & $71.3 \pm 1.1$ & $\bf{19.1 \pm 0.9}$ & $\bf{65.4 \pm 1.9}$ & - & - &  - & - \\
                    En{\textendash}Ar & $10.1 \pm 0.6$ & $75.3 \pm 1.3$ & $9.2 \pm 0.6$ & $77.2 \pm 0.9$ & $\bf{12.4 \pm 0.7}$ & $\bf{69.8 \pm 0.9}$ & - & - & - & - \\
                    \bottomrule
                \end{tabular}
            \end{minipage}}
            \caption{Experimental results comparing the translation quality produced by using the different tokenizers. In the columns \emph{Mecab} and \emph{Stanford}, \emph{Moses tokenizer} was used for segmenting the English part of the corpora since both \emph{Mecab} and \emph{Stanford Word Segmenter} only work for Japanese and Chinese respectively. Best results are denoted in bold.}
            \label{tabTok}
        \end{table}
        
        For the Ja{\textendash}En experiment, the best results were yielded by \emph{Moses tokenizer} and \emph{Mecab}. It must be taken into account that in both experiments, the English side of the corpus was segmented with \emph{Moses tokenizer}, this means that the segmentation of the target side has a greater impact on the translation quality. Overall, there is a quality improvement of around 4 points in terms of BLEU and 3 points in terms of TER with respect to the tokenizer which yielded the second best results.
        
        For En{\textendash}Ja, the best results were yielded by \emph{Mecab}, representing a significant improvement (around 12 points in terms of BLEU and 15 points in terms of TER) with respect to the tokenizer which yielded the second best results. Most likely, this is due to \emph{Mecab} being developed specifically to segment Japanese.
        
        For Ru{\textendash}En and En{\textendash}Ru, \emph{Moses tokenizer} yielded the best results (with improvements of around 2 to 4 points in terms of BLEU and 5 points in terms of TER). It is worth noting that, in both cases, \emph{SentencePiece} and \emph{OpenNMT tokenizer} yielded similar results.
        
        The Chinese experiments behaved similarly to the Japanese experiments: \emph{Moses tokenizer} and \emph{Stanford Word Segmenter} (the specific Chinese word tokenizer, which included using \emph{Moses tokenizer} for segmenting the English part of the corpus) achieved the best results when translating to English (yielding an improvement of around 7 points in terms of BLEU and 5 points in terms of TER), and \emph{Stanford Word Segmenter} achieved the best results when translating to Chinese (yielding an improvement of around 8 points in terms of BLEU and 20 points in terms of TER).
        
        For the German experiments, the best results were yielded by both \emph{OpenNMT tokenizer} and \emph{Moses tokenizer}, representing an improvement of around 7 to 9 points in terms of BLEU and 14 to 17 points in terms of TER. It is worth noting how, despite being the largest corpora, \emph{SentencePiece}{\textemdash}which learns how to segment from the corpora's training data{\textemdash}yielded the worst results. As a future study, we should evaluate the relation between the size of the corpora and the quality yielded by \emph{SentencePiece}.
        
        Finally, Arabic behaved similarly to Russian, with \emph{Moses tokenizer} yielding the best results for both Ar{\textendash}En and En{\textendash}Ar (representing improvements of around 2 to 4 points in terms of BLEU and 4 to 6 points in terms of TER). However, \emph{SentencePiece} performed similar to \emph{Moses tokenizer} when translating to English. When translating to Arabic, both \emph{SentencePiece} and \emph{OpenNMT tokenizer} yielded similar results.
        
        Overall, \emph{Moses tokenizer} yielded the best results for German, Russian and Arabic experiments. When using specialized morphologically oriented tokenizers, the system using \emph{Mecab} obtained the best results for Japanese experiments; and \emph{Stanford Word Segmenter} for Chinese experiments. Additionally, \emph{OpenNMT tokenizer} and \emph{SentencePiece} yielded the worst translation quality in all experiments. An explanation for these poor results is that \emph{OpenNMT tokenizer} is fairly simple: it only separates punctuation symbols from words. However, this is not the case for \emph{SentencePiece}. We think that using \emph{SentencePiece} in a bigger training dataset in order to better learn the segmentation could help to improve their results. Nonetheless, as mentioned before, we have to corroborate this in a future work.
        
    \section{Qualitative Analysis}
    \label{SecAnalysis}
    
    \begin{table} [!ht]
        \small
        \resizebox{0.83\textwidth}{!}{\begin{minipage}{\textwidth}
                \begin{tabular}{l|l }
                    \toprule
                    \multicolumn{2}{c}{\textbf{Example 1}}\\
                    \hline
                    \textbf{Source}	& Revalidation of single-pilot single-engine class ratings \\
                    \textbf{Reference} & verlängerung von klassenberechtigungen für einmotorige flugzeuge mit einem piloten \\
                    \textbf{SentencePiece} & \textbf{verlängerung} \emph{der einzelantriebsklasse einmotorischer motorklasse} \\ 
                    \textbf{OpenNMT tokenizer} & \emph{zur validierung der einmotorik-einzelmaschine} \textbf{mit} \emph{einzelantrieb} \\ 
                    \textbf{Moses tokenizer} & \textbf{verlängerung von klassenberechtigungen für einmotorige flugzeuge mit einem piloten} \\
                    \hline
                    \multicolumn{2}{c}{\textbf{Example 2}}\\
                    \hline
                    \textbf{Source}	& Cold drawing of wire \\
                    \textbf{Reference} & herstellung von kaltgezogenem draht \\
                    \textbf{SentencePiece} &	\emph{kalt zeichnung des drahtes} \\ 
                    \textbf{OpenNMT tokenizer}	& \emph{kaltbildzeichnung} \\ 
                    \textbf{Moses tokenizer} &	\textbf{herstellung von kaltgezogenem draht} \\ 
                    \bottomrule
                \end{tabular}
            \end{minipage}}
            \caption{English to German translation examples comparing \emph{SentencePiece}, \emph{OpenNMT tokenizer} and \emph{Moses tokenizer}. First line corresponds to the source sentence in English, second line to the German reference and third, forth and fifth lines to the translations generated using \emph{SentencePiece}, \emph{OpenNMT tokenizer} and \emph{Moses tokenizer} respectively to segment the corpora. Correct translations hypothesis are denoted in bold, and incorrect translations are denoted in italic.}
            \label{ta:analysis}
        \end{table}
        
        
        We obtained a better performance using \emph{Moses tokenizer} than \emph{OpenNMT tokenizer} and \emph{SentencePiece}. In order to qualitatively analyze this performance, 
        \cref{ta:analysis} shows a couple of examples of translation outputs generated using \emph{SentencePiece}, \emph{OpenNMT tokenizer} and \emph{Moses tokenizer} for segmenting the corpora.
        
        The first example clearly shows a better performance when using \emph{Moses tokenizer} rather than \emph{SentencePiece}. The translation output from the system trained using \emph{Moses tokenizer} for segmenting matches the reference. However, the output translations of the systems using \emph{OpenNMT tokenizer} and \emph{SentencePiece} are wrong. Translation segmented with \emph{OpenNMT tokenizer} contains many repetitions and lacks sense. Additionally, translation segmented by \emph{SentencePiece} has problems repeating some words in the translation (e.g., \emph{motor}) and missing some translation words (e.g., the translation of \emph{pilot}). 
        
        The system's behavior using \emph{Moses tokenizer} in the second example is similar: its translation matches the reference. By contrast, the systems using \emph{SentencePiece} and \emph{OpenNMT tokenizer} translated wrongly. The system using \emph{SentencePiece} translated all the words from the source but its translation is not grammatically correct. A correct translation could be \emph{kalte Zeichnung des Drahtes}. Lastly, \emph{OpenNMT tokenizer}'s performance is the worst in this case: the translation of its system ignored the word \emph{wire}.
        
        Therefore, we observed that, despite sharing the same data and model architecture, the behavior of the systems' translation changed as a result of using a different tokenizer.
        
    \section{Conclusions}
    \label{SecConclusion}
    In this study, we tested different tokenizers to evaluate their impact on the quality of the final translation. We experimented using 10 language pairs and arrived to the conclusion that tokenization has a great impact on the translation quality, achieving gains of up to 12 points of BLEU and 15 points of TER.
    
    Additionally, we observed that there was not a single best tokenizer. Each one produced the best results for certain language pairs. Although, in some cases, those best results overlapped with the ones yielded by other tokenizers. Moreover, we have seen different behaviors depending on the language pair direction. The system using \emph{SentencePiece} obtained the best results for Ar{\textendash}En, but not for En{\textendash}Ar translation.
    
    As a future work, we would like to evaluate the relation between the size of the corpora and the quality yielded by \emph{SentencePiece}{\textemdash}which uses each language's training corpora to learn how to segment. It would also be interesting to compare more segmentation strategies such as separating by characters or fixed n-grams. Finally, we would like to confirm that repeating these experiments on some of the general domain training data used for these languages achieves similar effects.

    \section{Acknowledgments}
    The research leading to these results has received funding from the Centro para el Desarrollo Tecnol{\'o}gico Industrial (CDTI) and the European Union through Programa Operativo de Crecimiento Inteligente (EXPEDIENT: IDI-20170964). We gratefully acknowledge the support of NVIDIA Corporation with the donation of a GPU used for part of this research.
    
    \bibliographystyle{apalike}
    \bibliography{arXiv}

\end{document}